%% file: main.tex
\documentclass[10pt,twocolumn,letterpaper]{article}

\usepackage{wacv}              


\definecolor{wacvblue}{rgb}{0.21,0.49,0.74}
\usepackage[pagebackref,breaklinks,colorlinks,allcolors=wacvblue]{hyperref}

\usepackage[accsupp]{axessibility}
\usepackage[T1]{fontenc}    
\usepackage{url}            
\usepackage{booktabs}       
\usepackage{amsfonts}       
\usepackage{nicefrac}       
\usepackage{microtype}      
\usepackage{xcolor}         
\usepackage{caption}
\usepackage{subcaption} %
\usepackage{graphicx}

\definecolor{mydarkblue}{rgb}{0.68, 0.85, 1.0}

\usepackage[utf8]{inputenc} 
\usepackage[T1]{fontenc}    

\usepackage{url}            
\usepackage{booktabs}       
\usepackage{amsfonts}       
\usepackage{nicefrac}       
\usepackage{microtype}      

\usepackage{sidecap}
\usepackage{float}
\usepackage{graphicx}

\usepackage{mathrsfs}

\usepackage{amsmath}
\usepackage{amssymb}
\usepackage{amsthm}
\usepackage{mathtools}
\usepackage[mathscr]{eucal}%
\usepackage{bm}
\usepackage{bbm}
\usepackage{relsize}
\usepackage{graphics}
\usepackage{wrapfig}
\usepackage{multirow}
\usepackage{enumitem}
\usepackage{tablefootnote}
\usepackage{makecell}

\usepackage{array}
\usepackage{color, colortbl}
\definecolor{Gray}{gray}{0.9}
\definecolor{darkpastelgreen}{rgb}{0.01, 0.75, 0.24}
\definecolor{cadetgrey}{rgb}{0.57, 0.64, 0.69}
\definecolor{camel}{rgb}{0.76, 0.6, 0.42}
\definecolor{lightskyblue}{rgb}{0.53, 0.81, 0.98}
\definecolor{lightsblue}{rgb}{0.53, 0.81, 0.98}
\definecolor{lightblue}{rgb}{0.68, 0.85, 0.9}
\definecolor{softblue}{rgb}{0.85, 0.91, 0.98}
\definecolor{mygray}{gray}{0.6}
\usepackage{algorithm}
\usepackage{algorithmic}
\usepackage{pgf}
\usepackage{xinttools}
\usepackage{tikz}
\usetikzlibrary{arrows.meta}
\usepackage{textcomp}
\usetikzlibrary{calc,shapes,arrows,positioning,automata,trees}
\usepackage{placeins} 
\usepackage{tabularx}
\usepackage{graphicx}
\usepackage{subcaption} 
\usepackage{listings}

\def\eg{\emph{e.g.}}

\def\etal{{\em et al.}}

\usepackage[font=small,labelfont=bf,tableposition=top]{caption}

\DeclareMathOperator*{\argmin}{arg\,min}

\newcommand{\ours}[0]{{QUOTA}\xspace}
\newcommand{\ourbench}[0]{{QUANT-Bench}\xspace}

\def\wacvPaperID{2221} 
\def\confName{WACV}
\def\confYear{2026}

\title{QUOTA: Quantifying Objects with Text-to-Image Models for Any Domain}

\author{
  Wenfang Sun\textsuperscript{1},
  Yingjun Du\textsuperscript{1},
  Gaowen Liu\textsuperscript{2},
  Yefeng Zheng\textsuperscript{3}\thanks{Corresponding author}~,
  Cees G. M. Snoek\textsuperscript{1} \\
  \textsuperscript{1}University of Amsterdam~
  \textsuperscript{2}Cisco Research~
  \textsuperscript{3}Westlake University~
}

\begin{document}
\maketitle
\input{sec/0_abstract}    
\input{sec/1_intro}

\input{sec/2_related}

\input{sec/3_preliminary}
\input{sec/4_methods}
\input{sec/5_experiments}
\input{sec/6_conclusions}


{
    \small
    \bibliographystyle{ieeenat_fullname}
    \bibliography{main}
}
\clearpage

\appendix

\input{sec/7_appendix}
\end{document}

%% file: sec/0_abstract.tex
\begin{abstract}
We tackle the problem of quantifying the number of objects by a generative text-to-image model. Rather than retraining such a model for each new image domain of interest, which leads to high computational costs and limited scalability, we are the first to consider this problem from a domain-agnostic perspective. We propose \ours, 
an optimization framework for text-to-image models that enables effective object quantification across unseen domains without retraining. It leverages a dual-loop meta-learning strategy to optimize a domain-invariant prompt. Further, by integrating prompt learning with learnable counting and domain tokens, our method captures stylistic variations and maintains accuracy, even for object classes not encountered during training.
For evaluation, we adopt a new benchmark \ourbench specifically designed for object quantification in domain generalization, enabling rigorous assessment of object quantification accuracy and adaptability across unseen domains in text-to-image generation.
Extensive experiments demonstrate that \ours outperforms conventional models in both object quantification accuracy and semantic consistency, setting a new benchmark for efficient and scalable text-to-image generation for any domain.

\end{abstract}

%% file: sec/1_intro.tex
\section{Introduction} 
\label{sec:intro}
\begin{figure}[t]
    \centering
\includegraphics[width=1.\linewidth]{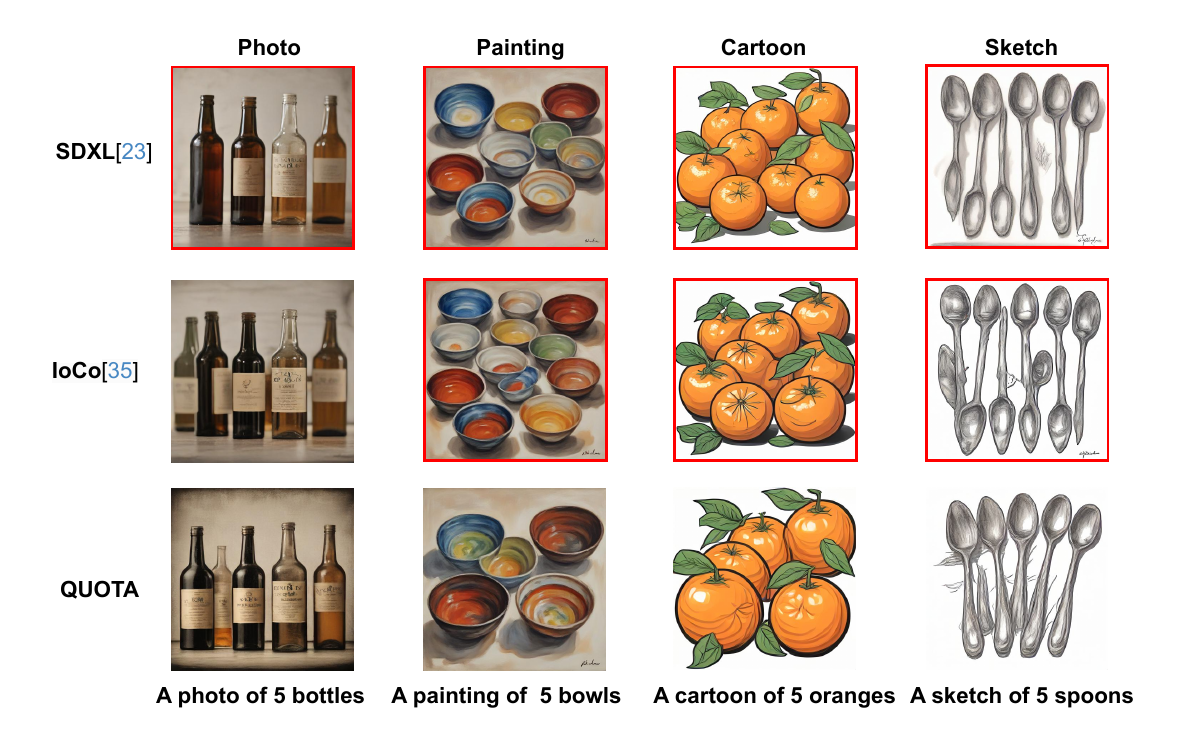} 
\caption{\textbf{Quantifying objects across domains.} In each column, one domain is held out as the unseen test domain (shown in the column title), while the remaining three domains are used for training. For example, in the first column \textit{Photo} is the test domain and the models are trained on \textit{Sketch}, \textit{Cartoon}, and \textit{Painting}. Red boxes indicate incorrect object counts. SDXL~\cite{sdxl} and IoCo~\cite{zafar2024iterative} struggle with consistent quantification across domains. In contrast, \ours accurately controls object counts across all domains, including unseen ones.}
\vspace{-4mm}

    \label{fig:intro}
\end{figure}

Text-to-image generative models~\cite{zhang2018learning, paiss2023teaching, starr2013number, dao2023flow} have achieved remarkable success in generating high-quality images from textual descriptions. However, accurately controlling the count of objects in these generated images remains challenging. Recently, Zafar \etal~\cite{zafar2024iterative} introduced prompt learning to refine the object counting prompts. While effective in controlled scenarios, this method struggles to maintain accurate object counts when confronted with varying domains. 
Similarly, CountGen~\cite{binyamin2025make} improves counting accuracy by guiding the denoising process through internal object representations; however, it is restricted to a single domain and depends on additional modules for layout prediction.
As a result, achieving domain-generalizable object quantification remains a crucial and unresolved challenge across various generative models, including diffusion and flow-matching models. Figure~\ref{fig:intro} illustrates this problem, showing how standard generative models like SDXL~\cite{sdxl} struggle with accurate object quantification across different domains, whereas our approach provides consistent and accurate results for any domain.

Domain generalization in machine learning aims to create models that perform well on unseen domains, reducing the need for retraining and enabling greater adaptability across diverse settings~\cite{zhou2022domain}. A wide range of approaches has been developed for this purpose, including methods that focus on learning domain-invariant features and leveraging diverse training environments~\cite{li2018learning, du2020learning, xiao2021bit}. Among these approaches, meta-learning has emerged as a promising technique, enabling models to adapt to new tasks by utilizing prior knowledge from related tasks~\cite{hochreiter2001learning, thrun2012learning, finn2017model}.  
Despite these advancements, there has been no exploration of meta-learning for domain-generalizable object quantification in text-to-image generation.  
In this paper, we fill this gap by proposing a novel meta-learning framework designed to enable robust object quantification across diverse and unseen domains within text-to-image models.

Specifically, we make the following three contributions in this paper. 
\textit{First}, we introduce a domain-agnostic setting for object quantification in generative text-to-image models, emphasizing the need for precise, domain-invariant object quantification. This setting is essential for adapting models to diverse and unseen prompts without retraining, addressing both scalability and efficiency challenges.
\textit{Second}, we propose a dual-loop meta-learning strategy for prompt optimization, designed to make prompts domain-invariant. In the training phase, we divide text prompts into meta-train and meta-validation domains. In the inner loop, we optimize prompt parameters on the meta-train domain to meet specific domain requirements. The outer loop refines these parameters on the meta-validation domain, allowing the model to learn a prompt that generalizes across unseen domains. This design mimics real-world scenarios where models must handle diverse domain shifts without prior exposure, ensuring adaptability to novel distributions in text-to-image generation.
\textit{Third}, we incorporate prompt learning with learnable quantification and domain tokens, enabling the model to capture stylistic variations effectively while preserving object quantification accuracy across domains. This approach allows for robust adaptation to new domains and unseen object classes without modifying the model architecture or retraining.

For evaluation, we adopt a new benchmark \ourbench specifically designed for object quantification in domain generalization, enabling rigorous assessment of accuracy and adaptability across unseen domains in text-to-image generation. Comparative evaluations against state-of-the-art methods demonstrate that \ours outperforms existing models in both object quantification accuracy and semantic alignment, setting a new benchmark for domain-generalizable text-to-image generation, enabling accurate object quantification across any domain without retraining.

%% file: sec/2_related.tex
\section{Related Work}
\label{sec: relatedwork}

\noindent\textbf{Text-to-image generation.} The field of image generation has experienced significant evolution, transitioning from initial Generative Adversarial Network (GAN)-based approaches~\cite{goodfellow2014generative} to the more recent diffusion-based models~\cite{ho2020denoising}. Early GAN methodologies laid the groundwork for generating images from textual descriptions, but often faced challenges related to training stability and output quality~\cite{goodfellow2014generative}. In contrast, diffusion models have emerged as a robust alternative, demonstrating enhanced performance in producing high-fidelity images conditioned on text prompts~\cite{ho2020denoising}. Notably, models like DALL·E 2~\cite{ramesh2022hierarchical} and Imagen~\cite{saharia2022photorealistic} have showcased the capability to generate photorealistic images from textual inputs. Concurrently, flow matching models have been introduced, offering advantages in training stability and sample quality by aligning continuous normalizing flows with target distributions~\cite{lipman2022flow}. Our approach builds upon these advancements in text-conditioned image generation, focusing on enhancing object counting consistency across diverse domains.

\noindent\textbf{Personalized image generation.}  
Personalized image generation aims to fine-tune models for specific concepts using limited examples. Textual Inversion~\cite{gal2022textual} learns a new token with just a few images, while DreamBooth~\cite{ruiz2022dreambooth} fine-tunes diffusion models for personalized content. However, adapting models for quantitative concepts, such as object counting, remains challenging as it requires precise numerical control rather than qualitative representation.
Recent works have improved prompt-image alignment through techniques like Prompt-to-Prompt~\cite{hertz2022prompt} for object manipulation, Null-inversion~\cite{mokady2023null} for real-image edits, and Attend-and-Excite~\cite{chefer2023attend} to ensure prompt consistency.
Our work introduces meta-optimized prompts that enable accurate, domain-generalizable object quantification without retraining, ensuring reliable control over object counts across diverse styles and contexts for consistent text-to-image generation.

\noindent\textbf{Object counting benchmarks.}
Several benchmarks have been proposed for evaluating object counting in text-to-image generation, including DrawBench\cite{saharia2022photorealistic}, CoCoCount\cite{binyamin2025make}, and YOLO-Count~\cite{zeng2025yolo}. DrawBench evaluates multiple aspects of image generation, including cardinality, but does not address domain generalization. CoCoCount focuses on counting accuracy within specific datasets (e.g., COCO), but remains domain-specific and does not generalize across unseen domains. On the other hand, YOLO-Count introduces benchmarks like OpenImg7-New and Obj365-New, which assess a model’s ability to generalize to unseen categories. In contrast, our \ours benchmark focuses on precisely generating objects across unseen domains without retraining, providing a more scalable solution for real-world applications.

\noindent\textbf{Domain generalization.}  
Domain generalization (DG) aims to enable models to generalize to unseen domains without access to target domain data during training. It is typically studied under two settings: multi-source DG, which leverages multiple source domains for robustness, and single-domain DG (SDG), which is more realistic but challenging due to limited data diversity.  Early DG approaches focused on domain-invariant feature learning via adversarial training~\cite{jia2020single,shao2019multi}, distribution alignment~\cite{li2018domain,li2020domain}, and self-supervised learning~\cite{carlucci2019domain,kim2021selfreg}. Meta-learning~\cite{du2020learning,shu2021open} has also been explored to enhance generalization.   However, no prior work has specifically addressed domain-generalizable object quantification in text-to-image diffusion models. We introduce a novel setting and methodology to achieve domain-invariant object quantification across diverse and unseen domains.

%% file: sec/3_preliminary.tex
\section{Preliminaries}  
\label{sec: preliminaries}

\noindent\textbf{Iterative object count optimization (IoCo).}
IoCo by Zafar \etal~\cite{zafar2024iterative} introduces an iterative method to optimize object count accuracy in text-to-image models. Given a generated image \( x \), an object category \( c \), and a target object count \( N \), their framework refines the generated output by minimizing the following objective:
\begin{equation}
\min_{\theta} \mathcal{L}_{\text{counting}}(x, c, N) + \lambda \mathcal{L}_{\text{semantic}}(x, c, N),
\end{equation}
where the counting loss ensures the estimated count aligns with the target:
\begin{equation}
\mathcal{L}_{\text{counting}}(x, c, N) = \left| g_{\text{cnt}}(x, c) - N \right|,
\end{equation}
with \( g_{\text{cnt}}(x, c) \) representing the estimated number of objects in the generated image. To maintain the semantic integrity of the image, a CLIP-based regularization is used:
\begin{equation}
\mathcal{L}_{\text{semantic}}(x, c, N) = \text{CLIP}(x, \text{``A photo of } N \text{ } c\text{''}).
\end{equation}
IoCo iteratively updates a learnable counting token \( e_{\text{cnt}} \) to adjust the model’s internal representation of quantity:
\begin{equation}
e_{\text{cnt}} \leftarrow e_{\text{cnt}} + \alpha \frac{\nabla_{e_{\text{cnt}}} \mathcal{L}(x, c, N)}{\|\mathcal{L}(x, c, N)\|_2}.
\end{equation}
%

\noindent\textbf{Meta-learning for domain generalization.}  
In this work, meta-learning aims to develop models capable of rapidly adapting to new tasks with limited data by leveraging knowledge from multiple tasks~\cite{finn2017model}. In the context of domain generalization, meta-learning seeks to train models that generalize well to unseen domains by simulating domain shifts during training~\cite{li2018learning}. Let $\mathcal{S}$ represent a training dataset comprising training and validation sets across various source domains $\mathcal{S}$, i.e., $\mathcal{S} = \{\mathcal{S} ^{\text{tr}}, \mathcal{S} ^{\text{val}}\}$, where \( \mathcal{S}^{\text{tr}} \) and \( \mathcal{S}^{\text{val}} \) are disjoint subsets of the source domain set \( \mathcal{S} \), with the former used for training and the latter for validation.
 The meta-learning process for domain generalization can be formulated as a bi-level optimization problem:
\begin{align}
\label{eq:bilevel-1}
    \min_{\phi}  &\; \sum_{s^{\text{val}} \in \mathcal{S}^{\text{val}}} \mathcal{L}_{\text{val}}({\theta_{s^{\text{val}}}^*}; s^{\text{val}}).     \\
\label{eq:bilevel-2}
    \text{s.t. } &\; {\theta_{s^{\text{val}}}^*}  = \argmin_{\theta_{s^{\text{tr}}}} \mathcal{L}_{\text{train}} ({\theta_{s^{\text{tr}}}}; s^{\text{tr}}), \quad s^{\text{tr}} \in \mathcal{S}^{\text{tr}},
\end{align}
where $\mathcal{L}_{\text{val}}$ and $\mathcal{L}_{\text{train}}$ denote the validation and training losses at the upper and lower levels, respectively. Here, ${\theta_{s^{\text{tr}}}}$ represents the domain-specific parameters for source domain $s^{\text{tr}}$, and $\phi$ denotes the shared meta-parameters. The lower-level optimization (Eq.~\eqref{eq:bilevel-2}) adapts $\theta$ to each source domain using the training set $\mathcal{S}^{\text{tr}}$ and the meta-parameters $\phi$. The upper-level optimization (Eq.~\eqref{eq:bilevel-1}) updates $\phi$ by minimizing the combined validation loss across source domains for the optimized domain-specific parameters ${\theta_{s^{\text{val}}}^*}$. During testing, the trained model is evaluated on data from unseen target domains $\mathcal{T}$, which were not accessible during training. The goal is to assess the model's ability to generalize to these novel domains without additional fine-tuning.


%% file: sec/4_methods.tex
\section{Methods}
\label{sec: methods}

In this section, we introduce quantifying objects with text-to-image models for any domain (\ours).
QUOTA leverages a dual-loop meta-learning framework to create domain-invariant prompts that guide object quantification without retraining, making it adaptable to new and unseen domains. Through prompt learning, QUOTA optimizes domain-specific and quantity-specific parameters to accurately adjust object quantities based on textual prompts, even when encountering new stylistic variations. Our framework operates on images generated by existing text-to-image models, such as SDXL~\cite{sdxl}, using object quantities inferred by a pre-trained detection model as feedback for optimization.

\subsection{Domain-generalizable object quantification}  

Accurately quantifying objects in text-to-image generative models is essential for ensuring alignment with specified numerical constraints. However, achieving consistent object counts across diverse domains is challenging due to the inherent variability in domain-specific styles and the lack of explicit numerical control in diffusion-based generation. 

Let $\mathcal{S}^{\text{tr}}$ represent a set of source domains used during training. Existing methods for object quantification in text-to-image diffusion models primarily focus on optimizing count accuracy within a fixed domain $s^{\text{tr}} \in \mathcal{S}^{\text{tr}}$ but overlook the challenge of generalization. These approaches assume that the model is trained and evaluated within the same domain, failing to account for domain shifts encountered when generating images in unseen distributions. As a result, a model trained on $\mathcal{S}^{\text{tr}}$ may overfit to domain-specific characteristics, leading to inconsistent quantification performance when applied to novel styles.

In this work, we formally introduce the problem of domain-generalizable object quantification in text-to-image diffusion models. Our goal is to ensure that object quantification remains reliable when applied to unseen target domains $\mathcal{T}$, where object distributions, styles, and scene compositions may differ considerably from $\mathcal{S}^{\text{tr}}$. To the best of our knowledge, this is the first study that investigates how to optimize object counting mechanisms in a way that remains robust across unseen domains. Without explicitly addressing domain generalization, object quantification methods risk becoming unreliable when applied to new domains, limiting their practical utility in real-world scenarios.

\begin{figure*}[t]
    \centering
\includegraphics[width=1.0\linewidth]{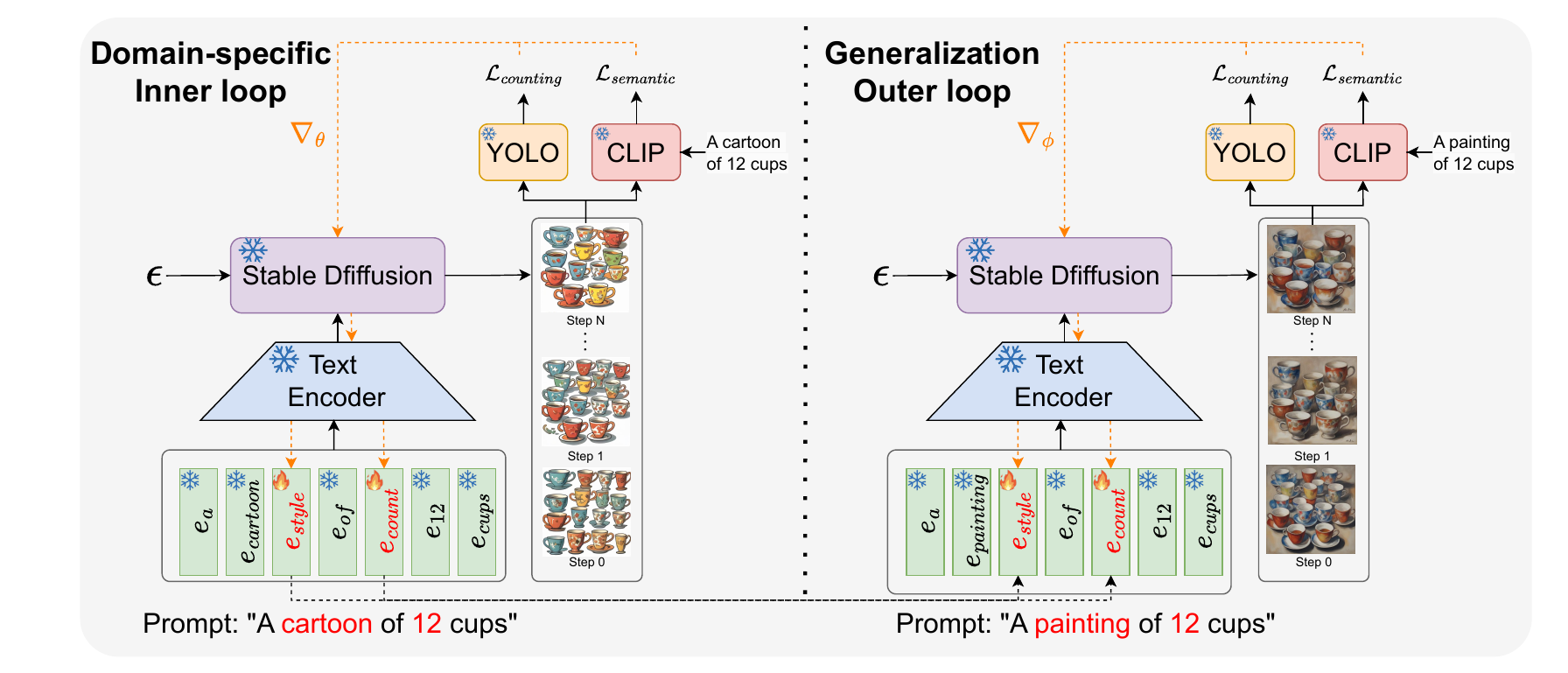}
\caption{\textbf{\ours's dual-loop meta-optimization.} The inner loop optimizes prompt parameters \( e_{\text{count}} \) and \( e_{\text{style}} \) across meta-train domains{ (\textit{Cartoon and Sketch})}  to adapt to domain-specific prompts, with losses computed using YOLOv9 for quantification accuracy and CLIP for semantic alignment.  The outer loop refines these parameters on a meta-validation domain {(\textit{Painting})}, enhancing the model’s generalization to unseen domains {(\textit{Photo})}. This dual-loop structure enables consistent object quantification and domain adaptation across diverse domains without retraining. }
\vspace{-4mm}
    \label{fig:framework}
\end{figure*}

\subsection{Dual-loop meta-optimization}  

\ours employs a dual-loop meta-optimization framework to achieve domain-invariant object quantification across diverse domains. This framework consists of an inner loop, which adapts the prompt parameters across multiple meta-train domains, and an outer loop, which refines these parameters on meta-validation domains to improve generalization. By dynamically sampling meta-train and meta-validation domains during training, the model learns to adapt to domain shifts and ensure robust object quantification. See Figure~\ref{fig:framework} for an overview.

\noindent\textbf{Meta-learning domain partitioning.} We denote the meta-train and meta-validation domains as $S^{tr}$ and $S^{val}$, respectively. During training, we dynamically partition $\mathcal{S}$ into meta-train domains $\mathcal{S}^{\text{tr}}$ and meta-validation domains $\mathcal{S}^{\text{val}}$. The model first optimizes domain-specific prompt parameters ${\theta_{s^{\text{tr}}}}$ using $\mathcal{S}^{\text{tr}}$, refining object quantification within each domain. It is then validated on $\mathcal{S}^{\text{val}}$ to refine domain-invariant prompt parameters $\phi$, promoting generalization beyond seen distributions.

\noindent\textbf{Inner loop optimization.}
In the inner loop, given a set of meta-train domains \( \mathcal{S}^{\text{tr}} \), we optimize the prompt parameters \( \theta \) across multiple domains to ensure the adaptability of the model. For each domain \( s^{\text{tr}} \in \mathcal{S}^{\text{tr}} \), the composite loss function includes: quantification loss $\mathcal{L}_{\text{counting}}$ and 
CLIP matching penalty $\mathcal{L}_{\text{semantic}}$. 

\textbf{Quantification loss.} 
Our inner loop optimization uses a differentiable counting function, \texttt{Count}, which estimates the quantity of objects for a specified class in an image. The quantification loss for a target class \( c \) is defined as the difference between the estimated and desired object counts, given by:
\begin{equation}
\mathcal{L}_{\text{counting}}(x_s^{\text{tr}}, c, N) = \left\| g_{\text{cnt}}(x_s^{\text{tr}}, c) - N \right\|,
\end{equation}
where $g_{\text{cnt}}(x_s^{\text{tr}}, c)$ provides an estimate of the object count for class \( c \) within the generated image \( x_s^{\text{tr}} \).
This loss is calculated using a detection-based dynamic scale, leveraging YOLOv9~\cite{yolov9} to estimate the count of objects of class \( c \) in the generated image \( x_s^{\text{tr}} \). The estimated count is given by:
\begin{equation}
    \text{Count}(x_s^{\text{tr}}, c) = \sum_{i=1}^{h \cdot w} \hat{\lambda}_{\text{scale}, x_s^{\text{tr}}, \mathcal{D}} \cdot \Phi_i(x_s^{\text{tr}}, c),
    \label{eq:counting_loss}
\end{equation}
where \( \hat{\lambda}_{\text{scale}, x_s^{\text{tr}}, \mathcal{D}} \) is a dynamically adjusted scaling factor for differentiation continuity, and \( \Phi_i(x_s^{\text{tr}}, c) \) represents the detection score for class \( c \) at the \( i \)-th bounding box~\cite{zafar2024iterative}. This loss penalizes the difference between the target count \( N \) and the detected count, ensuring accurate object quantification.

\textbf{CLIP matching penalty.} To preserve image semantics, we add a CLIP-based penalty term:
\begin{equation}
\mathcal{L}_{\text{semantic}}(x_s^{\text{tr}}, c, N) = \text{CLIP}(x_s^{\text{tr}}, \text{``A } s_d \text{ of } N \text{ } c \text{''}),
\label{eq:penalty_loss}
\end{equation}
where \( s_d \) is a domain descriptor specific to the domain  \( s \) (\eg, ``photo,'' ``sketch,'' ``cartoon,'' or ``painting''), and CLIP provides a matching score between the generated image \( x_s^{\text{tr}} \) and the textual description ``A [style] of N [class],'' reinforcing both semantic and stylistic alignment. The inner loop loss \( \mathcal{L}_{\text{inner}} \) is then computed by summing these losses across all meta-train domains:
\begin{align}
\mathcal{L}_{\text{inner}}({\theta_{s^{\text{tr}}}}) = 
\sum_{s^{\text{tr}} \in \mathcal{S}^{\text{tr}}}
\mathcal{L}_{\text{counting}}(x_s^{\text{tr}}, c, N; {\theta_{s^{\text{tr}}}})  \\
+ \lambda \mathcal{L}_{\text{semantic}}(x_s^{\text{tr}}, c, N; {\theta_{s^{\text{tr}}}}),
\label{eq:inner_loss}
\end{align}
where \( \lambda \) is a hyperparameter balancing the importance of object quantification accuracy and semantic consistency. This inner loop optimization updates the prompt parameters \( \theta \) to adapt to the specific characteristics of the meta-train domains.

\noindent\textbf{Outer loop optimization.}
The outer loop further refines the prompt parameters by evaluating them on the meta-validation domain \( \mathcal{S}^{\text{val}} \) to promote generalization to unseen domains. After optimizing the prompt parameters ${\theta_{s^{\text{val}}}^*}$ in the inner loop across meta-train domains, the outer loop computes the loss on the meta-validation domain, defined as:
\begin{align}
\mathcal{L}_{\text{outer}}(\phi) = 
\mathcal{L}_{\text{counting}}(x^{\text{val}}_s, c, N; {\theta_{s^{\text{val}}}^*}) \\
+ \lambda \mathcal{L}_{\text{semantic}}(x^{\text{val}}_s, c, N; {\theta_{s^{\text{val}}}^*}),
\label{eq:outer_loss}
\end{align}
where \( x^{\text{val}}_s \) is the generated image using the updated prompt parameters ${\theta_{s^{\text{val}}}^*}$ from the inner loop, and \(\phi\) denotes the shared meta-parameters, i.e., the global initialization of \(e_{\text{count}}\) and \(e_{\text{style}}\). 
The inner loop adapts these tokens to domain-specific requirements, while the outer loop updates their shared initialization to improve generalization across domains. Unlike the inner loop, where the text prompt follows the structure ``A \( s_d^{\text{tr}} \) of  N \( c \)'', the outer loop utilizes a different domain descriptor derived from \( \mathcal{S}^{\text{val}} \), modifying the prompt format to ``A \( s_d^{\text{val}} \) of N \( c \)''. This ensures that the model is exposed to different domain styles between training and validation, reinforcing robustness to domain shifts. By training on $\mathcal{S}^{\text{tr}}$ and validating on $\mathcal{S}^{\text{val}}$, the outer optimization step updates \( \phi \), ensuring that the learned prompt parameters generalize effectively to new, unseen domains. This process achieves accurate object quantification while maintaining semantic fidelity. Through this dual-loop meta-optimization framework, \ours adapts the prompt parameters to handle domain shifts effectively, enabling consistent object quantification across various domains without compromising image quality or semantic alignment.

\subsection{Learnable quantification and domain tokens}

To achieve adaptive control over both object quantification and the domain in text-to-image generation, we introduce prompt learning with two learnable tokens, \( e_{\text{count}} \) and \( e_{\text{style}} \). These tokens enhance the model's ability to generalize across diverse domains by encoding information specific to object quantification and stylistic features directly within the prompt. Notably, all other model parameters remain fixed, allowing the model to achieve domain adaptability without altering the underlying architecture.
To exercise control over the generated image, we concatenate two newly added pseudo token embeddings, \( e_{\text{count}} \) and \( e_{\text{style}} \), to the text embedding. These embeddings are randomly initialized and are iteratively updated during training to adaptively encode information about object quantification and domain. Unlike optimizing existing tokens, which may corrupt tokens with multiple meanings or associations, the introduction of these pseudo tokens allows the model to learn domain-specific adjustments without impacting predefined tokens. Formally, we optimize the embeddings \( e_{\text{count}} \) and \( e_{\text{style}} \) for each image generated, where the prompt parameters \( \theta \) in our meta-optimization framework refer to these learnable tokens.
In the inner loop, \( e_{\text{count}} \) and \( e_{\text{style}} \) are trained across meta-train domains to capture domain-specific prompt characteristics, resulting in domain-specific embeddings. The outer loop then refines these embeddings by adjusting the meta-parameters to generalize across domains, using the meta-validation domain to guide the final optimization. The outer loop’s updated meta-parameters represent our optimized, domain-generalizable prompt embeddings, ensuring consistency in object quantification and domain adaptation even for unseen domains.

\subsection{Testing phase}
During testing on a new domain $\mathcal{T}$, we simply incorporate the domain and quantification information from the test text prompt by concatenating it with our learned tokens, \( e_{\text{style}} \) and \( e_{\text{count}} \). For example, if the input text prompt is ``a cartoon of 6 apples,'' the prompt fed to SDXL~\cite{sdxl}  becomes ``a cartoon \( e_{\text{style}} \) of 6 \( e_{\text{count}} \) apples.'' This approach allows us to generate images accurately aligned with the desired domain and object count without needing to re-optimize \( e_{\text{style}} \) and \( e_{\text{count}} \), thus saving time and ensuring efficient adaptation to new domains.

%% file: sec/5_experiments.tex
\begin{table*}[!t]
\centering
\scalebox{0.62}{
\begin{tabular}
{lrrrrrrrrrrrrrrrr}
\toprule
& \multicolumn{3}{c}{Photo} & \multicolumn{3}{c}{Painting} & \multicolumn{3}{c}{Cartoon} & \multicolumn{3}{c}{Sketch} & \multicolumn{3}{c}{Average} \\
\cmidrule(lr){2-4} \cmidrule(lr){5-7} \cmidrule(lr){8-10} \cmidrule(lr){11-13} \cmidrule(lr){14-16}
& \multicolumn{2}{c}{YOLO} & CLIP-S $\uparrow$  & \multicolumn{2}{c}{YOLO} & CLIP-S $\uparrow$ & \multicolumn{2}{c}{YOLO} & CLIP-S $\uparrow$ & \multicolumn{2}{c}{YOLO} & CLIP-S $\uparrow$ & \multicolumn{2}{c}{YOLO} & CLIP-S $\uparrow$ \\
\cmidrule(lr){2-3} \cmidrule(lr){5-6} \cmidrule(lr){8-9} \cmidrule(lr){11-12} \cmidrule(lr){14-15}
& MAE $\downarrow$ & RMSE $\downarrow$ & & MAE $\downarrow$ & RMSE $\downarrow$ & & MAE $\downarrow$ & RMSE $\downarrow$ & & MAE $\downarrow$ & RMSE $\downarrow$ & & MAE $\downarrow$ & RMSE $\downarrow$ & \\
\midrule
w/o Meta-optimization & 12.75 & 16.75 & 73.2 & 10.40 & 13.78 & 74.0 & 11.49 & 13.28 &73.1 & 11.33 & 12.92 & 71.9 & 11.49 & 14.18 & 73.1 \\
\rowcolor{gray!20}
\textbf{w/ Meta-optimization}  & {\textbf{10.09}} & {\textbf{13.98}} & {\textbf{74.3}} & \textbf{8.04} & \textbf{11.04} & \textbf{75.9} & \textbf{9.48} & \textbf{11.92} & \textbf{74.8} & \textbf{9.15} & \textbf{11.69} & \textbf{74.1}  & \textbf{9.19} & \textbf{12.16} & \textbf{74.8} \\
\bottomrule
\end{tabular}}
\caption{\textbf{Benefit of meta-optimization} on object quantification and semantic alignment across different domains on \ourbench.  Meta-optimization shows improvements across all metrics, especially in challenging domains such as \textit{Sketch}, demonstrating its effectiveness in enhancing both object quantification and domain generalization.}
\label{ab_meta-optimization}
\end{table*}

\vspace{-4mm}

\begin{table*}[!t]
\centering
\scalebox{0.62}{
\begin{tabular}{lcccccccccccccccc}
\toprule
& \multicolumn{3}{c}{Photo} & \multicolumn{3}{c}{Painting} & \multicolumn{3}{c}{Cartoon} & \multicolumn{3}{c}{Sketch} & \multicolumn{3}{c}{Average} \\
\cmidrule(lr){2-4} \cmidrule(lr){5-7} \cmidrule(lr){8-10} \cmidrule(lr){11-13} \cmidrule(lr){14-16}
& \multicolumn{2}{c}{YOLO} & CLIP-S $\uparrow$  & \multicolumn{2}{c}{YOLO} & CLIP-S $\uparrow$ & \multicolumn{2}{c}{YOLO} & CLIP-S $\uparrow$ & \multicolumn{2}{c}{YOLO} & CLIP-S $\uparrow$ & \multicolumn{2}{c}{YOLO} & CLIP-S $\uparrow$ \\
\cmidrule(lr){2-3} \cmidrule(lr){5-6} \cmidrule(lr){8-9} \cmidrule(lr){11-12} \cmidrule(lr){14-15}
& MAE $\downarrow$ & RMSE $\downarrow$ & & MAE $\downarrow$ & RMSE $\downarrow$ & & MAE $\downarrow$ & RMSE $\downarrow$ & & MAE $\downarrow$ & RMSE $\downarrow$ & & MAE $\downarrow$ & RMSE $\downarrow$ & \\
\midrule
SDXL~\cite{sdxl} & 14.77 & 20.26 & 73.2 & 12.27 & 17.17 &73.5 & 14.10 & 17.97 & 72.5 & 11.92 & 16.51 & 71.4 & 13.26 & 17.98 & 72.7 \\
 w/ $e_{\text{style}}$ & 14.41 & 19.26 & 73.3 & 11.31 & 15.28 &73.7 & 12.73 & 15.07 &72.9 & 9.91 & 13.25 & 72.7 & 12.28 & 16.15 &  73.2\\
w/ $e_{\text{count}}$&  14.12& 19.60 & 73.9 & 10.20 &  14.44&  74.1 & 10.66 & 13.17 & 73.7 & 8.75 & 11.89 &  73.3 &10.94 & 14.76 & 73.8 \\
\rowcolor{gray!20}
\textbf{ Ours (w/ $e_{\text{count}}$ \& $e_{\text{style}}$)}  & {\textbf{10.09}} & {\textbf{13.98}} & {\textbf{74.3}} & \textbf{8.04} & \textbf{11.04} & \textbf{75.9} & \textbf{9.48} & \textbf{11.92} & \textbf{74.8} & \textbf{9.15} & \textbf{11.69} & \textbf{74.1}  & \textbf{9.19} & \textbf{12.16} & \textbf{74.8} \\
\bottomrule
\end{tabular}}
\caption{\textbf{Benefit of learnable tokens} on object quantification and semantic alignment across domains on \ourbench. Adding learnable tokens \( e_{\text{count}} \) and \( e_{\text{style}} \) shows improvements across all metrics, with the best performance achieved when both tokens are used together. }
\vspace{-6mm}
\label{ab_tokens}
\end{table*}

\section{Experiments}
\label{sec: experiment}

\subsection{Experimental setup}
\noindent\textbf{QUANT-Bench.}  
We developed \textit{QUANT-Bench}  to assess object quantification performance across various classes, quantities, and visual domains. The classes in QUANT-Bench are based on the FSC dataset by Ranjan \etal~\cite{ranjan2021learning}, a widely used dataset for object counting that includes 19 classes such as common objects, vehicles, and animals. For each class, we generated 25 examples for quantities ranging from one to 25, yielding a total of 475 samples. 
To evaluate the model’s ability to generalize across different domains, we created images in four distinct domains: \textit{Photo}, \textit{Painting}, \textit{Cartoon}, and \textit{Sketch}. Both training and testing classes are drawn from the FSC dataset. To test each domain, we used a leave-one-domain-out setting, where two domains are used as meta-train domains for the inner loop, one domain is used as the meta-validation domain for the outer loop, and the remaining domain is held out as the unseen test domain. This setup allows us to measure how well the model generalizes object quantification to unseen domains by training with textual descriptions from the other three domains.

\noindent\textbf{Metrics.}  
To evaluate quantification accuracy, we use the Mean Absolute Error (MAE) and the Root Mean Squared Error (RMSE) based on YOLOv9~\cite{yolov9}. These metrics are calculated by comparing the detected object counts to the target count in each image, with MAE capturing the average absolute deviation and RMSE highlighting larger errors through squared deviations. {For semantic alignment assessment, we use the CLIP-S from Eq.~\eqref{eq:penalty_loss}, following~\cite{hessel2021clipscore}. Conventional metrics such as FID~\cite{heusel2017gans} and Inception Score~\cite{salimans2016improved} are not appropriate for our \ourbench. FID requires a real reference distribution, which is not available in our synthetic prompt space, and the Inception Score emphasizes sample diversity rather than alignment with the input prompt.}

\noindent\textbf{Implementation details.}  
We conducted training and evaluation on a single NVIDIA A6000 GPU with 48GB of memory. Training each token requires approximately two minutes, totaling around 30 iterations, using the SDXL framework~\cite{sdxl}. For image quality, we find a single denoising step suffices. The optimized quantification token can be reused without additional optimization time. We set \( \lambda {=} 5 \), a learning rate of 0.01 for optimization, and the scaling hyperparameter \( \lambda_{\text{scale}} \) to 60 for a static scale. Code will be released.

\subsection{Results}

\begin{figure}[t]
    \centering
    \includegraphics[width=1.0 \linewidth]{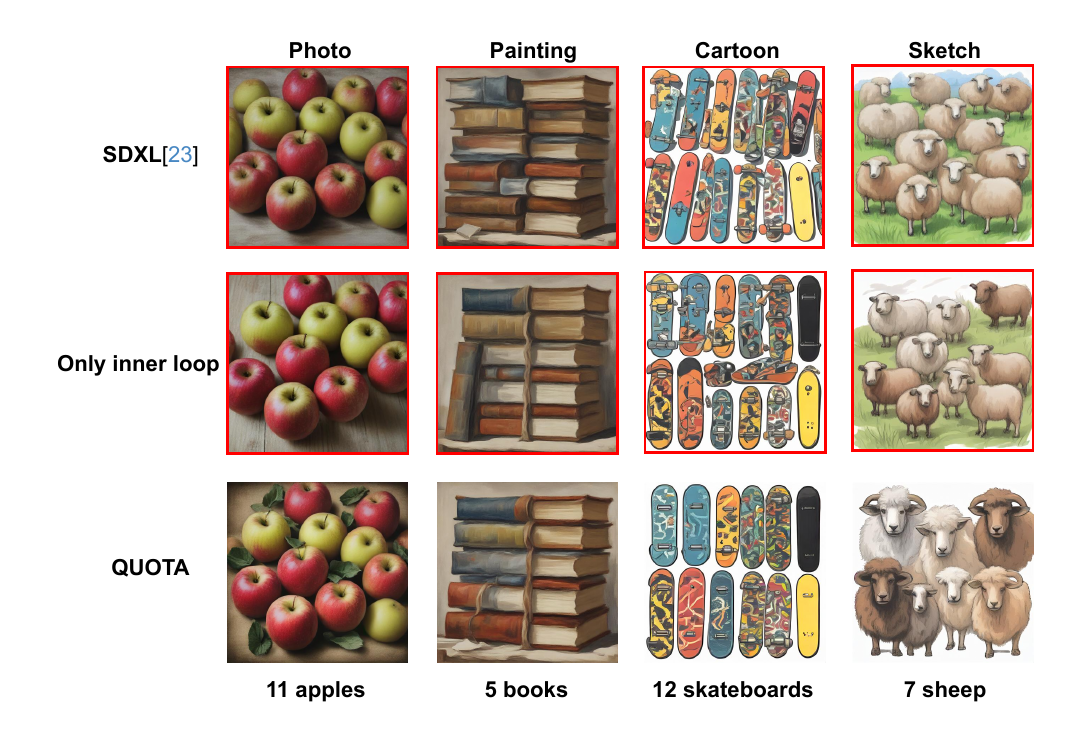}
\caption{\textbf{Benefit of meta-optimization.} We compare SDXL~\cite{sdxl}, our method with only the inner loop, and full \ours. Without meta-optimization, models struggle with accurate quantification. \ours achieves consistent counting across domains.}
\vspace{-4mm}
    \label{fig:inner}
\end{figure}

\noindent\textbf{Benefit of meta-optimization.}  
Table~\ref{ab_meta-optimization} demonstrates the impact of meta-optimization on object quantification accuracy and semantic alignment across different image styles. On average, meta-optimization achieves substantial improvements in both MAE and RMSE, highlighting enhanced object quantification accuracy and adaptability. The benefits are particularly pronounced in challenging domains, such as \textit{Cartoon}, where meta-optimization leads to improved precision and consistency. Additionally, meta-optimization improves semantic alignment, as indicated by higher CLIP-S scores, reflecting closer alignment with prompt semantics.
Figure~\ref{fig:inner} provides qualitative examples, comparing outputs from stable diffusion, an inner-loop-only optimization, and our full \ours framework. While inner-loop-only optimization offers some gains, it still struggles with consistent counts. In contrast, our complete meta-optimization approach, which includes both inner and outer loops, enables precise and domain-invariant object quantification across all styles, even in difficult cases like \textit{Sketch}. These results confirm that meta-optimization considerably enhances the model's ability to generalize quantification and semantic consistency across diverse domains.

\noindent\textbf{Benefit of learnable tokens.}  
Table~\ref{ab_tokens} demonstrates the impact of adding learnable tokens \( e_{\text{count}} \) and \( e_{\text{style}} \) on object quantification accuracy and semantic alignment across different domains. The standard SDXL, without any learnable tokens, shows limited adaptability, achieving lower accuracy across all domains. Introducing the \( e_{\text{style}} \) token alone provides moderate improvements, indicating that adapting to domain style contributes positively to performance. 
Adding only the \( e_{\text{count}} \) token leads to further enhancements, emphasizing the importance of object count adaptation. When both learnable tokens \( e_{\text{count}} \) and \( e_{\text{style}} \) are combined, as in \ours, the model achieves the best performance across all metrics and domains.
These results confirm that the joint use of \( e_{\text{count}} \) and \( e_{\text{style}} \) strengthens the model’s ability to generalize object quantification across diverse domains while maintaining robust semantic alignment.

\begin{figure}[t]
    \centering
    \includegraphics[width=1.0\linewidth]{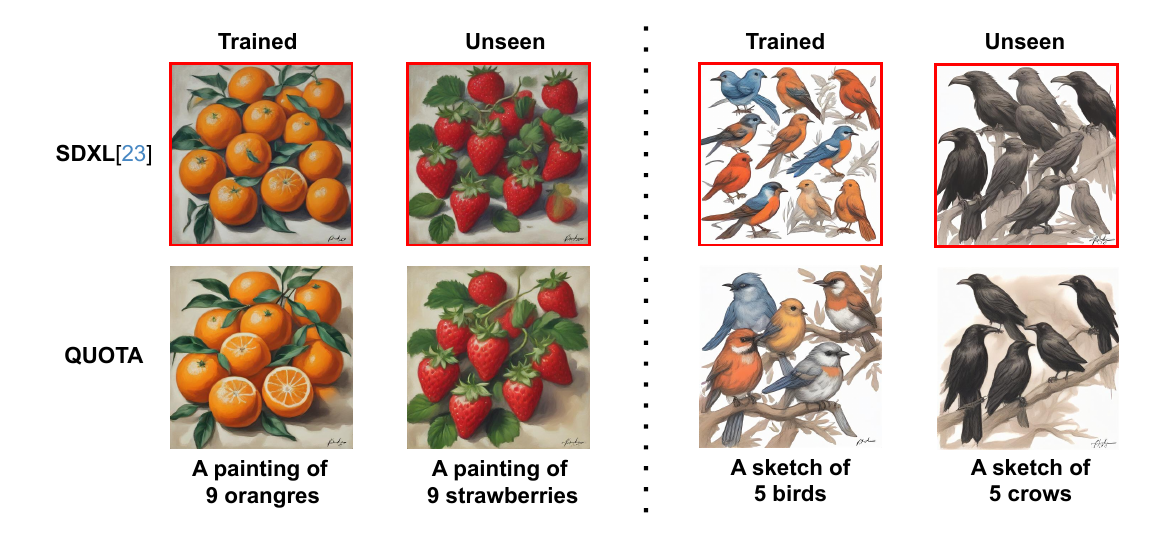}
\caption{\textbf{Generalization to unseen classes.} Comparison of SDXL~\cite{sdxl} and \ours on both seen and unseen classes. \ours maintains accurate quantification and style consistency, while SDXL struggles with unseen classes, highlighting our method’s robustness.}    \label{fig:generalization_unseen_classes}
\end{figure}

\begin{table*}[t]
\centering
\scalebox{0.62}{
\begin{tabular}
{lrrrrrrrrrrrrrrrr}
\toprule
& \multicolumn{3}{c}{Photo} & \multicolumn{3}{c}{Painting} & \multicolumn{3}{c}{Cartoon} & \multicolumn{3}{c}{Sketch} & \multicolumn{3}{c}{Average} \\
\cmidrule(lr){2-4} \cmidrule(lr){5-7} \cmidrule(lr){8-10} \cmidrule(lr){11-13} \cmidrule(lr){14-16}
& \multicolumn{2}{c}{YOLO} & CLIP-S $\uparrow$  & \multicolumn{2}{c}{YOLO} & CLIP-S $\uparrow$ & \multicolumn{2}{c}{YOLO} & CLIP-S $\uparrow$ & \multicolumn{2}{c}{YOLO} & CLIP-S $\uparrow$ & \multicolumn{2}{c}{YOLO} & CLIP-S $\uparrow$ \\
\cmidrule(lr){2-3} \cmidrule(lr){5-6} \cmidrule(lr){8-9} \cmidrule(lr){11-12} \cmidrule(lr){14-15}

& MAE $\downarrow$ & RMSE $\downarrow$ & & MAE $\downarrow$ & RMSE $\downarrow$ & & MAE $\downarrow$ & RMSE $\downarrow$ & & MAE $\downarrow$ & RMSE $\downarrow$ & & MAE $\downarrow$ & RMSE $\downarrow$ & \\
\midrule
SDXL~\cite{sdxl} & 14.77 & 20.26 & 73.2 & 12.27 & 17.17 &73.5 & 14.10 & 17.97 & 72.5 & 11.92 & 16.51 & 71.4 & 13.26 & 17.98 & 72.7 \\
Only train domains & 12.75 & 16.75 & 73.2 & 10.40 & 13.78 & 74.0 & 11.49 & 13.28 &73.1 & 11.33 & 12.92 & 71.9 & 11.49 & 14.18 & 73.1 \\
IoCo~\cite{zafar2024iterative} & \textbf{7.72} &\textbf{12.21} & \textbf{74.4} & 11.36 & 15.68 & 73.1 & 12.40 & 15.93 &72.7 & 10.80 & 14.21 & 71.9 & 10.57 & 14.51 & 73.0\\ 
\midrule
\rowcolor{gray!20}
\textbf{\ours}   & {10.09} & {13.98} & {74.3} & \textbf{8.04} & \textbf{11.04} & \textbf{75.9} & \textbf{9.48} & \textbf{11.92} & \textbf{74.8} & \textbf{9.15} & \textbf{11.69} & \textbf{74.1}  & \textbf{9.19} & \textbf{12.16} & \textbf{74.8} \\
\midrule
\textcolor{gray!75} {Target Domain} & \textcolor{gray!75}{5.68} & \textcolor{gray!75}{8.63} & \textcolor{gray!75}{75.9} & \textcolor{gray!75}{4.61} & \textcolor{gray!75}{7.12} &  \textcolor{gray!75}{76.8} & \textcolor{gray!75}{5.31} & \textcolor{gray!75}{7.86} & \textcolor{gray!75}{75.9}& \textcolor{gray!75}{5.94} & \textcolor{gray!75}{8.53} & \textcolor{gray!75}{75.3} & \textcolor{gray!75}{5.38} & \textcolor{gray!75}{8.04} & \textcolor{gray!75}{76.0} \\
\textcolor{gray!75}{Train \& Target Domain} & \textcolor{gray!75}{6.43} & \textcolor{gray!75}{9.05} & \textcolor{gray!75}{74.7} & \textcolor{gray!75}{7.61} & \textcolor{gray!75}{10.79} & \textcolor{gray!75}{76.1}  & \textcolor{gray!75}{8.43} & \textcolor{gray!75}{10.96} & \textcolor{gray!75}{75.1} & \textcolor{gray!75}{8.75} & \textcolor{gray!75}{11.27} & \textcolor{gray!75}{74.7} & \textcolor{gray!75}{7.81} & \textcolor{gray!75}{10.52} & \textcolor{gray!75}{75.2}\\
\bottomrule
\end{tabular}}
\caption{\textbf{Comparison with others} on object quantification and semantic alignment across domains in \ourbench. \textcolor{gray!75}{Gray} numbers denote target-domain upper bounds. While IoCo~\cite{zafar2024iterative} is trained on \textit{Photo}, \ours performs well across all domains without such supervision, highlighting its generalizability.}
\label{tab:ab_comparison_baselines}
\end{table*}

\begin{table*}[t]
\centering
\scalebox{0.65}{
\begin{tabular}{lcccccccccccccccc}
\toprule
& \multicolumn{3}{c}{Photo} & \multicolumn{3}{c}{Painting} & \multicolumn{3}{c}{Cartoon} & \multicolumn{3}{c}{Sketch} & \multicolumn{3}{c}{Average} \\
\cmidrule(lr){2-4} \cmidrule(lr){5-7} \cmidrule(lr){8-10} \cmidrule(lr){11-13} \cmidrule(lr){14-16}
& \multicolumn{2}{c}{YOLO} & CLIP-S $\uparrow$  & \multicolumn{2}{c}{YOLO} & CLIP-S $\uparrow$ & \multicolumn{2}{c}{YOLO} & CLIP-S $\uparrow$ & \multicolumn{2}{c}{YOLO} & CLIP-S $\uparrow$ & \multicolumn{2}{c}{YOLO} & CLIP-S $\uparrow$ \\
\cmidrule(lr){2-3} \cmidrule(lr){5-6} \cmidrule(lr){8-9} \cmidrule(lr){11-12} \cmidrule(lr){14-15}
& MAE $\downarrow$ & RMSE $\downarrow$ & & MAE $\downarrow$ & RMSE $\downarrow$ & & MAE $\downarrow$ & RMSE $\downarrow$ & & MAE $\downarrow$ & RMSE $\downarrow$ & & MAE $\downarrow$ & RMSE $\downarrow$ & \\
\midrule
SDXL~\cite{sdxl} & 14.07 & 19.80 & 72.8 & 20.64 & 27.05 & 73.0 & 11.71 & 15.40 & 72.1 & 18.39 & 24.03 & 71.1 & 16.20 & 21.57 & 72.3 \\

IoCo~\cite{zafar2024iterative} &  \textbf{13.42} & \textbf{18.97} & \textbf{73.3} & 13.16 & 19.01&  73.4 & 11.29 & 14.73 & 72.7 & 14.41 & 19.71 &  71.60 & 13.07& 18.11 &  72.8\\

\rowcolor{gray!20}
\textbf{\ours}    & 13.72 & 19.20 & 73.1 & \textbf{10.99} & \textbf{15.24} & \textbf{74.5} & \textbf{9.94} & \textbf{12.22} & \textbf{73.2} & \textbf{12.05} & \textbf{14.68} & \textbf{72.3}  & \textbf{11.68} & \textbf{15.33} & \textbf{73.3} \\
\bottomrule
\end{tabular}}

\caption{\textbf{Generalization to unseen classes.} 
Comparison of SDXL~\cite{sdxl}, IoCo~\cite{zafar2024iterative}, and \ours on unseen classes across four target domains. 
\ours achieves consistently lower MAE and RMSE together with higher CLIP-S scores, demonstrating stronger quantification accuracy, semantic alignment, and robustness to domain shifts.}
\vspace{-8mm}
\label{tab:unseen}
\end{table*}

\begin{figure*}[h]
    \centering
\includegraphics[width=1.0\linewidth]{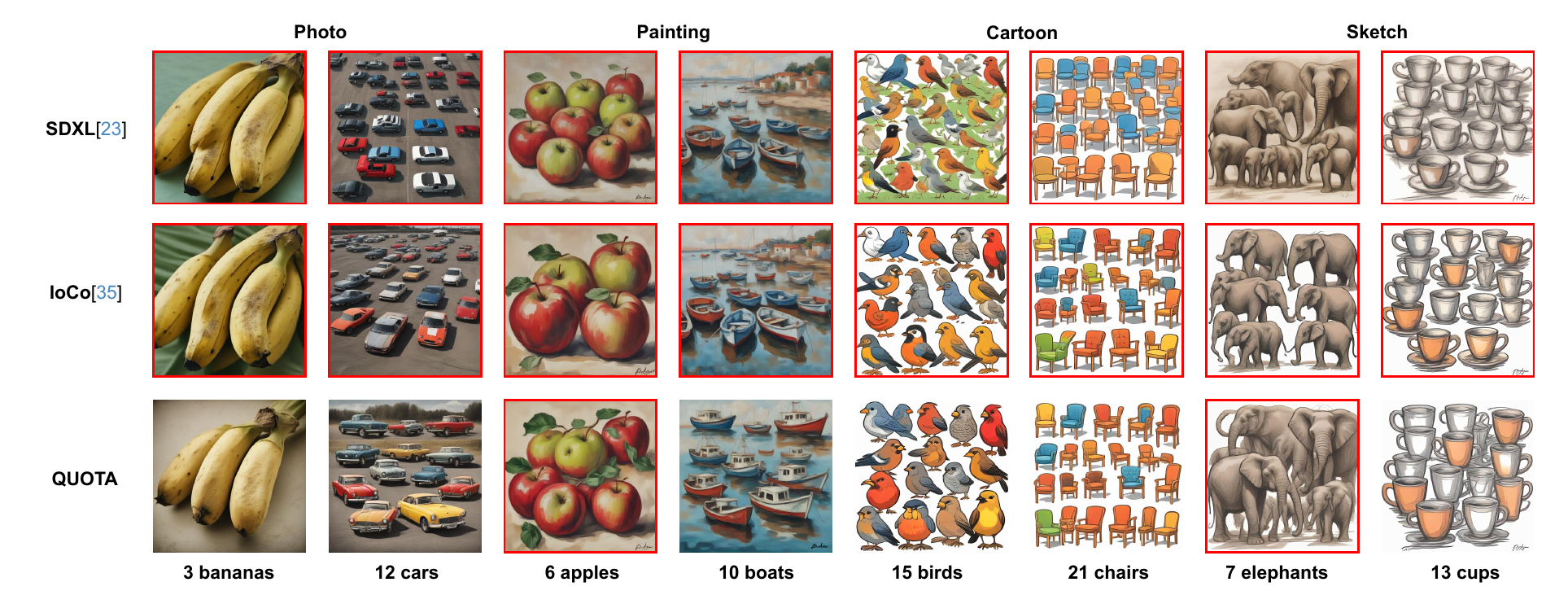}
\vspace{-4mm}
\caption{\textbf{Comparison with others} by their generated images across different domains and object classes. This figure highlights the quantification accuracy and domain consistency of \ours. In the Painting domain (apples), QUOTA generates seven apples, but one appears significantly smaller, making it difficult for YOLOv9 to detect. Similarly, in the Sketch domain, the highly stylized and intricate setting challenges accurate object quantification, illustrating the limitations of detection-based counting in complex visual styles.}
 \vspace{-6mm}
    \label{fig:overall_performance}
\end{figure*}

\noindent\textbf{Generalization to unseen classes.}  
We further evaluate the ability of \ours, trained on 19 predefined classes, to generalize to classes not seen during training. Figure~\ref{fig:generalization_unseen_classes} provides qualitative examples, where we test on new categories such as \textit{strawberries} and \textit{crows}. While SDXL~\cite{sdxl} struggles to maintain correct counts and style consistency, \ours successfully adapts to these unseen categories and domains, producing accurate quantification and faithful semantic alignment. Table~\ref{tab:unseen} complements this analysis with quantitative results across four target domains. The unseen classes are grouped by semantic similarity to training classes (e.g., apples generalize to tomatoes, oranges, and strawberries; birds to crows, pigeons, and seagulls; sheep to zebras, horses, and cows). We report MAE, RMSE, and CLIP-S scores averaged over nine unseen classes with quantities from 1 to 25. Across all domains (\textit{Photo}, \textit{Painting}, \textit{Cartoon}, and \textit{Sketch}), \ours consistently outperforms SDXL~\cite{sdxl} and IoCo~\cite{zafar2024iterative}, achieving substantially lower MAE and RMSE together with higher CLIP-S scores. These qualitative and quantitative results together highlight \ours's robustness in extending object quantification to unseen classes while preserving semantic consistency, demonstrating its broad applicability across diverse domains and distributions.


\begin{table}[t]
\centering
    \scalebox{1.0}{
      \begin{tabular}{lrrr}
\toprule
{Methods}  & {MAE $\downarrow$}  & {RMSE $\downarrow$} & {CLIP-S $\uparrow$} \\
\midrule
 Simple   & 9.19 & 12.16 & 74.8   \\
 Complex  & 7.23  & 9.01 &  73.1  \\    
 \midrule
Simple $\rightarrow$ complex & 8.25  & 10.05 & 73.8   \\ 
\bottomrule
\end{tabular}}
\caption{\textbf{Effect of prompt complexity.} \textit{Simple} uses default prompts, while \textit{Complex} includes GPT-4o-generated descriptions and trains the count token. Applying tokens from \textit{Simple} to \textit{Complex} (\textit{Simple $\rightarrow$ Complex}) shows our method's robustness to prompt variation.}
\vspace{-6mm}
\label{tab:prompt_complexity}
\end{table}

\noindent\textbf{Effect of prompt complexity.} To evaluate the impact of prompt complexity on object quantification, we compare three settings: Simple, Complex, and Simple $\rightarrow$ Complex, as shown in Table~\ref{tab:prompt_complexity}. The 
Simple setting uses default prompts, while Complex incorporates GPT-4o-generated~\cite{OpenAI_GPT4_2023} detailed descriptions and trains the count token as a learnable token, leading to improved performance.
To assess the adaptability of our method, we apply the style and count tokens trained on simple prompts directly to complex prompts, which results in performance of the "Simple to Complex" setting. This demonstrates our method effectively generalizes across prompt complexities, maintaining robust object quantification without requiring additional retraining.

\noindent\textbf{Comparison with others.}  
To provide a comprehensive understanding of \ours's performance, we include a qualitative comparison in Figure~\ref{fig:overall_performance} alongside the quantitative results in Table~\ref{tab:ab_comparison_baselines}. Figure~\ref{fig:overall_performance} illustrates generated images across various domains and classes, with \ours demonstrating superior quantification accuracy and domain consistency. In contrast, both SDXL~\cite{sdxl} and IoCo~\cite{zafar2024iterative} struggle with incorrect object quantification across different domains, as indicated by the red outlines on images with quantification errors. \ours achieves consistent object counts matching the target quantities in most cases, while the baseline models frequently misestimate counts, especially in complex domains. Notably, IoCo's reliance on training within the photo domain explains its higher performance in this domain but highlights its limitations in generalizing to unseen styles. 
Together, these visual and numerical comparisons underscore \ours's capability to set a new benchmark for object quantification accuracy and domain adaptability in text-to-image generation.


%% file: sec/6_conclusions.tex
\section{Conclusion}
\label{sec: conclusion}

We presented \ours, a novel optimization framework for domain-agnostic object quantification in text-to-image generation. Unlike existing methods that require retraining for each domain, \ours enables accurate quantification across diverse and unseen domains via a dual-loop meta-learning strategy that optimizes a domain-invariant prompt. By introducing learnable quantification and domain tokens into the prompt space, our approach effectively captures stylistic variations and maintains accuracy—even on object classes not seen during training. To facilitate systematic evaluation, we proposed a new benchmark tailored for domain-generalizable object quantification, enabling rigorous assessment of model adaptability and precision. Extensive experiments demonstrated that \ours outperforms standard baselines in both count accuracy and semantic alignment, establishing a new standard for flexible and scalable text-to-image generation.

\noindent\textbf{Limitations.}
Despite its strong cross-domain performance, \ours faces challenges in highly stylized or abstract domains where object boundaries are ambiguous. The dual-loop meta-learning introduces additional training overhead due to iterative prompt optimization, though test-time efficiency remains higher than adaptation-based methods such as IoCo~\cite{zafar2024iterative}. Furthermore, our reliance on external object detectors may introduce counting errors. Prior work~\cite{paiss2023teaching} has shown that models like CLIP struggle with numerical reasoning, indicating that integrating counting-aware modules could further improve accuracy. Future work will explore stronger object detectors and extend \ours to other generative models, enabling applications in fine-grained attribute control and compositional scene generation.

\section*{Disclaimer}
Funded by the European Union. Views and opinions expressed are however those of the author(s) only and do not necessarily reflect those of the European Union or the European Commission. Neither the European Union nor the European Commission can be held responsible for them.

\section*{Acknowledgments}
\noindent This was supported by European Union’s Horizon Europe research and innovation programme under grant agreement number 101214398 (ELLIOT).

%% file: sec/7_appendix.tex
\begin{table*}[!t]
    \centering
    \scalebox{0.8}{
    \begin{tabular}{c ccccc ccccc ccc}
     \midrule
    \textbf{Models} &
    \multirow{1}{*}{{\textbf{1}}}&
    \multirow{1}{*}{{\textbf{2}}}&
    \multirow{1}{*}{{\textbf{3}}}&
    \multirow{1}{*}{{\textbf{4}}}&
    \multirow{1}{*}{{\textbf{5}}}&
    \multirow{1}{*}{{\textbf{6}}}&
    \multirow{1}{*}{{\textbf{7}}}&
    \multirow{1}{*}{{\textbf{8}}}&
    \multirow{1}{*}{{\textbf{9}}}&
    \multirow{1}{*}{{\textbf{10}}}&
    \multirow{1}{*}{{\textbf{11}}}&
    \multirow{1}{*}{{\textbf{12}}}&
    \multirow{1}{*}{{\textbf{13}}}\\
    \midrule
    SDXL~\cite{sdxl} & 3.80 & \textbf{1.13} & \textbf{2.93} & \textbf{3.27} & 6.47 & 8.53 & 8.07 & 13.40 & 12.80 & 16.53 & 17.73 & 16.63 & 17.60 \\
    IoCo~\cite{zafar2024iterative}  & 7.20 & 1.93 & 3.20 & 3.60 &  6.40& 8.00 & 9.33 & 12.60 & 12.00 & 14.27 &  16.13 & 12.67  & 16.80\\
    \rowcolor{gray!20}
    \textbf{\ours} & \textbf{2.80} & 1.53 & 3.40 & 5.27 & \textbf{6.33} & \textbf{5.73} & \textbf{7.27} & \textbf{8.13} & \textbf{9.47} & \textbf{13.13} & \textbf{9.60}& \textbf{10.27} & \textbf{10.40}\\
    \midrule
    \textbf{Models} &
    \multirow{1}{*}{{\textbf{14}}}&
    \multirow{1}{*}{{\textbf{15}}}&
    \multirow{1}{*}{{\textbf{16}}}&
    \multirow{1}{*}{{\textbf{17}}}&
    \multirow{1}{*}{{\textbf{18}}}&
    \multirow{1}{*}{{\textbf{19}}}&
    \multirow{1}{*}{{\textbf{20}}}&
    \multirow{1}{*}{{\textbf{21}}}&
    \multirow{1}{*}{{\textbf{22}}}&
    \multirow{1}{*}{{\textbf{23}}}&
    \multirow{1}{*}{{\textbf{24}}}&
    \multirow{1}{*}{{\textbf{25}}}&
    \multirow{1}{*}{\textbf{Avg}}\\
    \midrule
    SDXL~\cite{sdxl} & 18.00 & 18.53 & 17.27& 17.47 & 17.00 & 20.20 & 18.80 & 17.20 & 18.13 & 17.93 & 18.87 & 24.47 & 14.10\\
    IoCo~\cite{zafar2024iterative}  &  \textbf{14.27} & 15.87 & 17.40 &  18.20& 16.27 &12.87  &  17.07& \textbf{12.73}  &\textbf{13.87}  & 14.73 & 14.47  & 18.13 & 12.40\\
    \rowcolor{gray!20}
    \textbf{\ours} & 15.00 & \textbf{10.00} & \textbf{11.87} & \textbf{8.80} & \textbf{7.80} & \textbf{9.13} & \textbf{13.87} & 13.40& 15.60 & \textbf{11.13} & \textbf{12.67} & \textbf{14.60} & \textbf{9.48} \\
    \bottomrule
    \end{tabular}}
    
\caption{\textbf{MAE comparison across 19 classes in the \textit{cartoon} target domain for varying object counts.} This table evaluates SDXL~\cite{sdxl}, IoCo~\cite{zafar2024iterative}, and \ours on object quantification for the object count ranging from 1 to 25, averaged across 19 classes in the \textit{cartoon}  target domain. \ours achieves the best performance in 19 out of 25 quantities, highlighting its superior accuracy and robustness in this target domain.}

\vspace{-2mm}
\label{tab:specialists_cartoon}
\end{table*}

\begin{table*}[!t]
    \centering
    \scalebox{0.8}{
    \begin{tabular}{c ccccc ccccc ccc}
     \midrule
    \textbf{Models} &
    \multirow{1}{*}{{\textbf{1}}}&
    \multirow{1}{*}{{\textbf{2}}}&
    \multirow{1}{*}{{\textbf{3}}}&
    \multirow{1}{*}{{\textbf{4}}}&
    \multirow{1}{*}{{\textbf{5}}}&
    \multirow{1}{*}{{\textbf{6}}}&
    \multirow{1}{*}{{\textbf{7}}}&
    \multirow{1}{*}{{\textbf{8}}}&
    \multirow{1}{*}{{\textbf{9}}}&
    \multirow{1}{*}{{\textbf{10}}}&
    \multirow{1}{*}{{\textbf{11}}}&
    \multirow{1}{*}{{\textbf{12}}}&
    \multirow{1}{*}{{\textbf{13}}}\\
    \midrule
    SDXL~\cite{sdxl} & 2.33 & \textbf{1.47} & 3.20 & 5.33 & 5.60 & 8.33 & 8.6 & 9.73 & 11.33 & 15.33 & 14.00 & 16.60 & 14.53 \\
    IoCo~\cite{zafar2024iterative}  & 3.67 & 1.53 & \textbf{2.40} & 3.93 &  6.73& 7.93 & 9.07 & 9.77 & 8.53 & 14.33 &  15.00 &11.67  & 12.67 \\
    \rowcolor{gray!20}

    \textbf{\ours} & \textbf{2.20} & 2.00 & 3.00 & \textbf{3.67} & \textbf{4.40} & \textbf{6.01} & \textbf{6.80} & \textbf{9.07} & \textbf{6.47} & \textbf{7.00} & \textbf{6.47}& \textbf{7.33} & \textbf{8.13}\\
    \midrule
    \textbf{Models} &
    \multirow{1}{*}{{\textbf{14}}}&
    \multirow{1}{*}{{\textbf{15}}}&
    \multirow{1}{*}{{\textbf{16}}}&
    \multirow{1}{*}{{\textbf{17}}}&
    \multirow{1}{*}{{\textbf{18}}}&
    \multirow{1}{*}{{\textbf{19}}}&
    \multirow{1}{*}{{\textbf{20}}}&
    \multirow{1}{*}{{\textbf{21}}}&
    \multirow{1}{*}{{\textbf{22}}}&
    \multirow{1}{*}{{\textbf{23}}}&
    \multirow{1}{*}{{\textbf{24}}}&
    \multirow{1}{*}{{\textbf{25}}}&
    \multirow{1}{*}{\textbf{Avg}}\\
    \midrule
    SDXL~\cite{sdxl} & 17.33 & 17.87 & 16.00 & 17.67 & 15.13 & 16.67 & 11.93 & 15.00 & 13.13 & 15.80 & 16.93 & 17.00 & 12.27\\
    IoCo~\cite{zafar2024iterative}  &  15.07& 17.40 & 14.67 &  14.80& 15.47 &15.87  &  13.40&15.27  &\textbf{12.93}  & \textbf{12.87} & 15.80  & \textbf{13.33} & 11.36\\
    \rowcolor{gray!20}
    \textbf{\ours} & \textbf{8.87} & \textbf{9.60} & \textbf{7.80} & \textbf{7
    .47} & \textbf{9.27} & \textbf{9.67} & \textbf{10.47} & \textbf{12.47}& 14.20 & 14.13 & \textbf{10.53} & 14.07 & \textbf{8.04} \\
    \bottomrule
    \end{tabular}}
    
\caption{\textbf{MAE comparison across 19 classes in the \textit{Painting} target domain for varying object counts.} This table evaluates SDXL~\cite{sdxl}, IoCo~\cite{zafar2024iterative}, and \ours on object quantification for the object count ranging from 1 to 25, averaged across 19 classes in the \textit{Painting} target domain. \ours achieves the best performance in 20 out of 25 quantities, highlighting its superior accuracy and robustness in this target domain.}

\vspace{-2mm}
\label{tab:specialists_painting}
\end{table*}

\section{Performance Across Object Quantities}
\begin{figure}[h]
    \centering
    \includegraphics[width=1.0\linewidth]{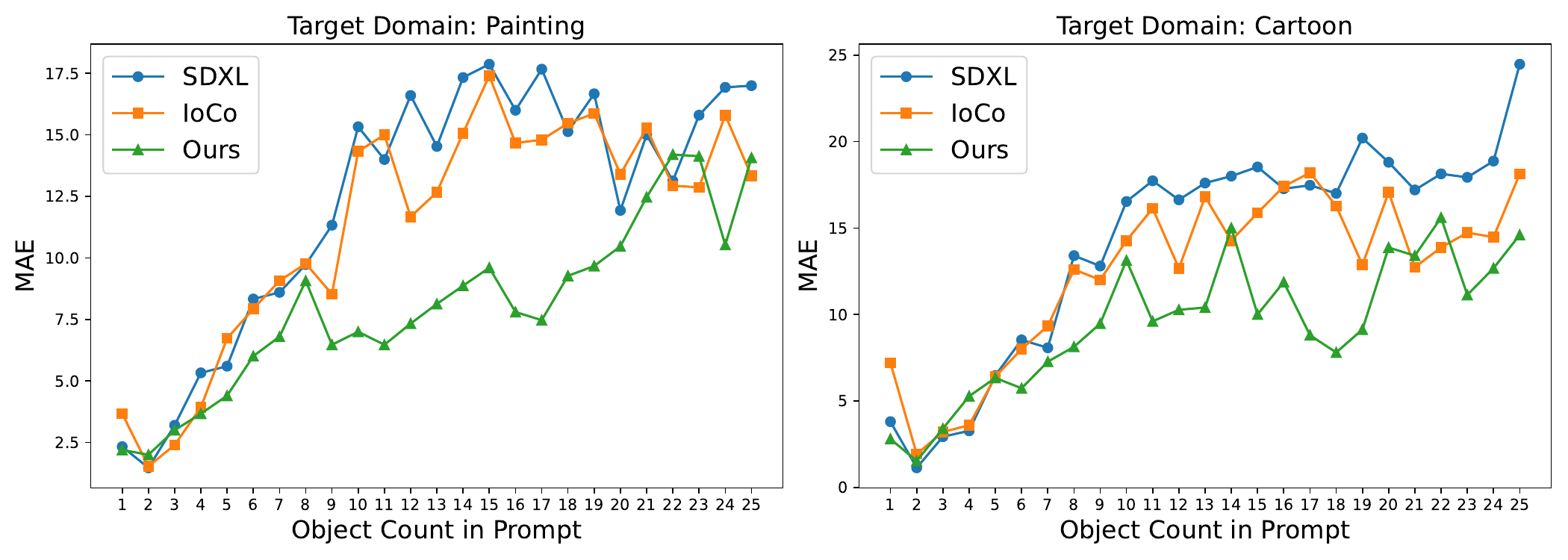}
    \caption{\textbf{MAE curves across object quantities for \textit{Painting} and \textit{Cartoon} domains.} 
    \ours significantly reduces error compared to SDXL~\cite{sdxl} and IoCo~\cite{zafar2024iterative}, especially in the common range of 5–20 objects.}
    \label{fig:mae_curve}
\end{figure}
Table~\ref{tab:specialists_painting} and Table~\ref{tab:specialists_cartoon} present the MAE results for 19 classes across two target domains: \textit{Painting} and \textit{Cartoon}, covering object quantities from 1 to 25. The results provide a detailed comparison between SDXL~\cite{sdxl}, IoCo~\cite{zafar2024iterative}, and \ours.

In the \textit{Painting} domain, \ours achieves the lowest MAE in 20 out of 25 quantities, consistently outperforming both IoCo~\cite{zafar2024iterative} and SDXL~\cite{sdxl}. Similarly, in the \textit{Cartoon} domain, \ours ranks first in 19 out of 25 cases, demonstrating its robustness in counting objects within stylized visual contexts. These findings highlight the strong performance of \ours in capturing complex visual variations and maintaining accurate object quantification across diverse classes and quantity ranges, further validating its effectiveness in real-world applications requiring precise object counting.

{Figure~\ref{fig:mae_curve} complements these tables by visualizing the per-count MAE curves for the two target domains. The curves reveal that \ours's largest gains concentrate in the practically most frequent range of 5--20 objects. In this interval, our method reduces the MAE by an average of 42.4\% over SDXL~\cite{sdxl} and 38.3\% over IoCo~\cite{zafar2024iterative} in the \textit{Painting} domain, and by 36.0\% and 28.8\% respectively in the \textit{Cartoon} domain. Below five objects, all three models already achieve very low errors, offering limited potential for further gains. In contrast, for counts above 20, the MAE remains high for all methods, likely due to increased visual complexity. These results indicate that \ours is particularly effective in the count range that matters most for real-world applications, while maintaining comparable performance on simpler cases.}


\section{Robust Generalization to Unseen Classes}

\begin{figure}[t]
    \centering
    \includegraphics[width=1.\linewidth]{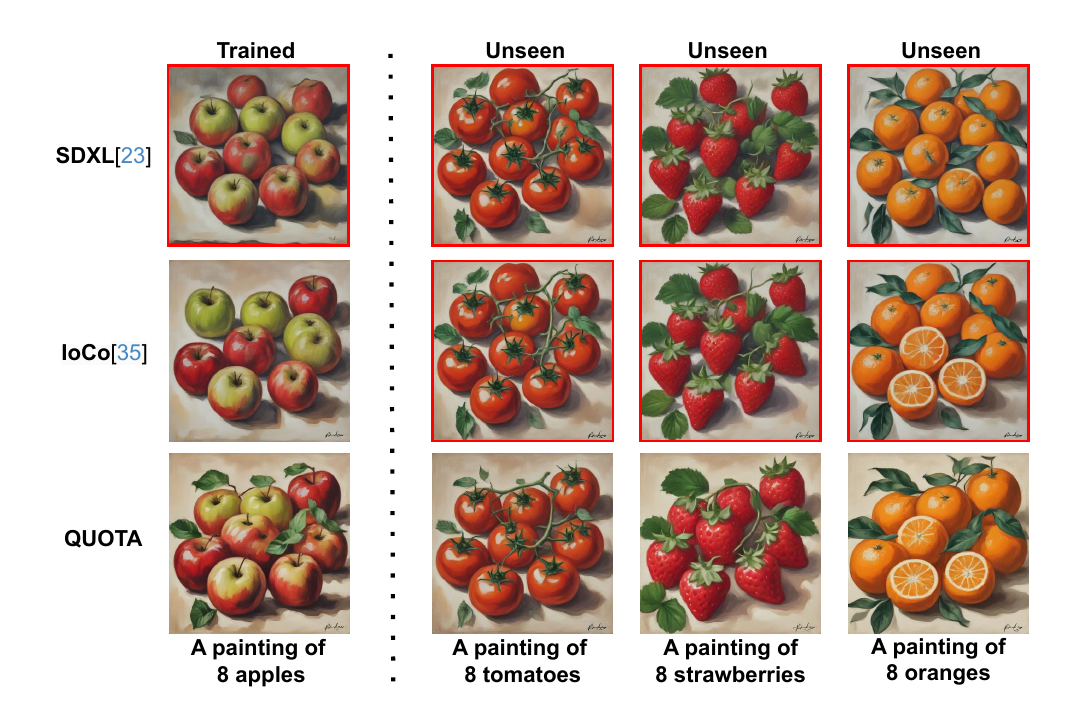}
\caption{\textbf{Generalization to unseen classes.} Comparison of object quantification between  SDXL~\cite{sdxl}, IoCo~\cite{zafar2024iterative}, and our \ours, evaluated on both trained and unseen  classes.} 
    \label{fig:unseen_classes}
\end{figure}

As shown in Figure~\ref{fig:unseen_classes}, we train the models using the 'apples' class in the painting target domain and evaluate their generalization on unseen classes with semantic similarity: 'tomatoes,' 'oranges,' and 'strawberries.' While IoCo~\cite{zafar2024iterative} performs well on the seen class during training, it fails to generalize effectively to unseen classes, leading to inaccurate quantification across all tested classes. Similarly, SDXL~\cite{sdxl} struggles to adapt, producing inconsistent object counts and exhibiting misalignment with textual prompts.

In contrast, our method, \ours, exhibits strong generalization, with accurate quantification across unseen classes. This highlights \ours’s ability to adapt to unseen classes and preserve accuracy, even across stylistically distinct domains. These results highlight the effectiveness of \ours in addressing the challenges of cross-domain and cross-class generalization.

